# TensorBank:
# Tensor Lakehouse for Foundation Model Training

Romeo Kienzler[1]
0000-0002-7145-8225

Leonardo Pondian Tizzei[2]
ltizzei@br.ibm.com

Benedikt Blumenstiel[1]
benedikt.blumenstiel@ibm.com

Zoltan Arnold Nagy[1]
nag@zurich.ibm.com

S. Karthik Mukkavilli[1]
karthik.mukkavilli@ibm.com

Johannes Schmude[2]
johannes.schmude@ibm.com

Marcus Freitag[2]
mfreitag@us.ibm.com

Michael Behrendt[3]
michaelbehrendt@de.ibm.com

Daniel Salles Civitarese[2]
sallesd@br.ibm.com

Naomi Simumba[2]
naomi.simumba1@ibm.com

Daiki Kimura[2]
daiki@jp.ibm.com

Hendrik Hamann[2]
hendrik@us.ibm.com

[1]IBM Research Europe
[2]IBM Research
[3]IBM Research and Development

*Abstract*—Storing and streaming high dimensional data for foundation model training became a critical requirement with the rise of foundation models beyond natural language. In this paper we introduce TensorBank – a petabyte scale tensor lakehouse capable of streaming tensors from Cloud Object Store (COS) to GPU memory at wire speed based on complex relational queries. We use Hierarchical Statistical Indices (HSI) for query acceleration. Our architecture allows to directly address tensors on block level using HTTP range reads. Once in GPU memory, data can be transformed using PyTorch transforms. We provide a generic PyTorch dataset type with a corresponding dataset factory translating relational queries and requested transformations as an instance. By making use of the HSI, irrelevant blocks can be skipped without reading them as those indices contain statistics on their content at different hierarchical resolution levels. This is an opinionated architecture powered by open standards and making heavy use of open-source technology. Although, hardened for production use using geospatial-temporal data, this architecture generalizes to other use cases like computer vision, computational neuroscience, biological sequence analysis and more.

## I. INTRODUCTION

Big data processing evolved from JBOD [1] systems like HDFS [2] and MapReduce [3] over Apache Spark [4] away from JBOD towards real-time query engines like Presto [5] or Apache Impala [6] on top of Cloud Object Storage (COS), which evolved as generic term for S3 compatible object storage. Storage formats evolved from CSV over JSON to high performance column stores like Parquet [7] and Avro [8], managed by data lakehouse storage managers like Apache Iceberg, Delta Lake or Apache Hudi [9]. Key driver for this transition is the tremendous increase of network performance with 400 Gbit/s Ethernet [10] becoming state of the art. Therefore, disks for big data processing were able to move out from the physical nodes which allowed for usage of cost-effective and elastic COS or dedicated storage systems on HPC clusters. These developments lead to the de-facto reference architecture shown in Figure 1. Although this architecture features nearly infinite scalability and relational database features like ACID [9], it is limited to tabular data with SQL as the only cross-cutting open standard interface, focusing on data discovery tasks, mainly composed of aggregations and filtering.

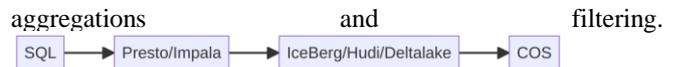

Figure 1: State of the art data lakehouse architecture

Foundation modeling entails building pre-trained transformers [11] with big data from multiple modalities in addition to natural language, that can be adapted to perform a range of downstream tasks [12]. Training large scale foundation models imposes different requirements on the big data architecture [13]. Firstly, data is organized in tensors and not in tables. Moreover, there are no aggregation operators involved. Filtering, on the other hand can become more complex as filter criteria often depend on tensor content. For example, feeding only cloud free Earth observation (EO) image tiles to models is more complex than just applying a filter criterion on individual floating-point numbers. In addition to growing volumes of EO data from next-generation satellites [14], increasingly gridded weather and climate model simulation outputs [15] are also being emulated with deep learning methods using tensors. There are efforts underway to utilize pre-trained transformers [16], with a vision to build Earth system Foundation models. Petabyte scale EO, weather and climate foundation modeling, thus necessitates a tensor lakehouse, storage and streaming service. In this work, we introduce TensorBank, the first tensor lakehouse for foundation model training on high dimensional datasets.

## II. ASSUMPTIONS AND GOALS

Our objective is to facilitate self-supervised (pre-) training, on large high-dimensional datasets as arise in the context of weather and climate, earth observation and medicine. This data is a series of images stacked along additional dimensions e.g., time, altitude, spectral band, ... and it should come as no surprise that application of techniques originally developed in the context of computer vision is prevalent. However, this data is the almost raw output of some sensing platform, medical device or simulation system. Both for this reason as well as concerns of scale, it is generally not directly fed to a deep learning algorithm. Instead, one typically identifies subtensors that are then loaded to GPU. These could comprise random samples, but also more complex subtensors. One might want to think of training a model for agriculture for which one requires subsets satellite images measuring 224x224 pixels that are free of clouds yet show high amounts of vegetation. In a climate or weather context [16] one wants to make sure that the training data fully reflects all possible precipitation conditions – whether extreme amounts or rare types like hail or freezing rain. Naively one might simply want to use all the available data for training. Yet the scales appearing in this context are so large that it is useful to define meaningful subsets. ESA's Sentinel 2 satellites generate about 4 PiB / year [14]. The entire ERA5 re-analysis comprises ~10 PB. In addition, as the examples show, one might define subsets based on different criteria. Thus, it is helpful to do subsetting as far as feasible without loading the raw data. Which is made more complicated by the fact that the criteria how to select subtensors often depends on the data itself. With all this said, we can state our overall goal: Given a large, high-dimensional dataset, we are need to dynamically identify semantically meaningful subtensor and effectively load them to GPU memory.

### III. ARCHITECTURE

#### A. Super tensors allowing for data addressing

As shown in Figure 2, TensorBank follows the generic data lakehouse pattern as depicted in Figure 1, but swapping and adding some components. First of all, the storage subsystem remains the same. Therefore, TensorBank can run in parallel to existing data lakehouse systems on the same cloud and HPC infrastructure. We are swapping out Apache Parquet and Apache Avro with ZARR [17].

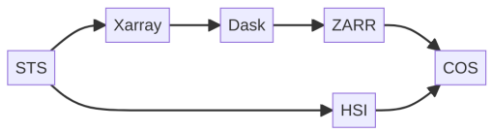

Figure 2: TensorBank component architecture

ZARR is a very interesting data format for storing and retrieving tensor data. What parquet and Avro is for tabular data, ZARR is for multidimensional tensor data. A ZARR folder stores tensor data of arbitrary dimensions. It keeps track of its contents using metadata situated at the root level of a ZARR folder. Actual data is organized in chunks, which is a predefined sub-tensor of the super tensor stored in a particular ZARR folder. This is how ZARR refers to data partitions. It is important to notice that ZARR supports sparse tensors as well. As the super tensor shape is defined in the metadata, chunk data is simply non-existent in the sparse case. This layout allows for minimizing number of reads for obtaining a particular scalar or sub-tensor from a ZARR folder as the (cached) metadata can be queried to obtain the file name containing a particular chunk of interest and within a chunk, the actual block can be obtained by computing the byte offset using the tensor indices used to address the data item.

Xarray [17] – similar to pandas for tabular data – is an interface layer which can read and write a variety of tensor data formats like netCDF [17], GeoTiff [18], Cloud Optimized GeoTiff [19] and ZARR. It allows for slicing and dicing through high dimensional tensor space using integer indices on the dimensions. In addition, for every dimension, integer indices are mapped to domain specific indices. This allows for advanced slicing and dicing using relational like query semantics by e.g., filtering on time intervals or coordinates.

Processing ZARR folders using Xarray often becomes resource intensive. Therefore, Xarray allows for delegating computation on chunks to DASK [17], which we use as a central query execution engine. When running on top of DASK, Xarray internally uses a DASK-array to parallelize processing on chunk level.

Up to this point, each ZARR folder on storage resembles a set of super-tensors containing a database or collection of data. Each scalar and every arbitrary sub-tensor shape is directly addressable via index. In case of ZARR on COS, ZARR's metadata is used to obtain chunk file names and also byte offsets within those chunk files. By using the HTTP range read functionality available on COS endpoints based on index and tensor-shape the exact byte arrays can be requested directly from COS.

#### B. Xarray for element indexing and filtering

As of now, in TensorBank, each tensor can be addressed by index. Because of its importance, we re-elaborate on Xarray's capability to mapped integer indices to an underlying domain specific value. This improves human readability and interpretability. Here are some examples:

- In four dimensional tensors representing tiles of EO imagery dimensions are labeled with longitude, latitude, timestamp and spectral band channel.
- Four dimensional tensors representing atmospheric data with dimensions: longitude, latitude, timestamp and height
- Similarly, in fMRI data dimensions are x, y, z, time - capturing changes in brain activity over the course of the scan.
- In biological sequence data, dimensions are sequence, position, feature, time, sample, and position.

Again, in summary, Xarray supports slicing and dicing, even using the underlying, domain specific index data types using relational algebra like semantics. The only thing Xarray can't do is, evaluating queries based on tensor content.

#### C. Filtering of tensors using Hierarchicl Staistical Indices

As shown above, filtering data by index is possible with Xarray out of the box. But often, content of sub-tensors decides if they should be part of a query result. This is true for explorative usage but also for (foundation) model training as certain data leads to poisoning of the model. Examples of content relevance for data selection include clouds on EO imagery, bacterial DNA in human tissue DNA samples, missing values or measurement errors. It is key to understand that removing these tensors can't be part of a data preprocessing step since their detection algorithm is often dependent on the downstream task and can be considered as part of the hyper-parameter set.

To avoid reading of a tensor just to decide on its removal, we introduce Hierarchical Statistical Indices (HSI). Based on the idea of overview layers introduced in the patent [20], in an indexing phase, sub-tensors of different hierarchical resolutions are grouped together and a set of domain specific statistics are computed. Here are some examples:

- In atmospheric data, at different resolutions, data points in a three-dimensional cube are grouped together and summary statistics like minimum, maximum and average temperature are calculated and stored to the HSI. This way, tensors containing temperature values below or above a certain threshold are not considered.

- In Earth Observation (EO), by application of a model, population density and land type are estimated and stored to the HSI at different resolutions. This way, only tensors containing non water and areas with low population density are identified.
- In fMRI data, four dimensional sub-tensors containing spatial coordinates x, y, z, and time are used to compute rate of change and are added to the HSI. This way, only tensors with high rate of change can be identified.

As can be seen in Figure 3, for the two-dimensional case, scalar values *a* to *bl* are grouped into four potential samples. Let us assume that these values correspond to pixels of EO imagery. Then, various domain specific and non-specific statistics are computed for each quadrant.

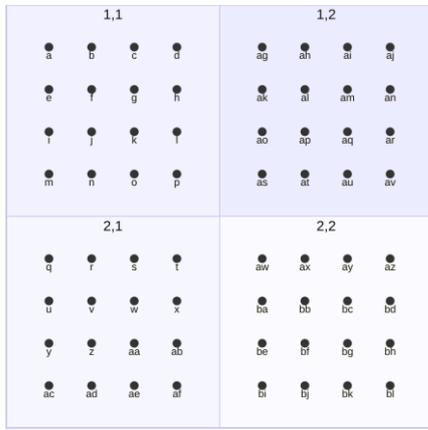

Figure 3: Grid layout of HSI

Figure 4 exemplifies a corresponding sub-tensor of the Level-4 HSI for the example above. The Level-4 HSI refers to reducing the resolution of the HSI by a factor of four on each dimension. Per grid cell, domain unspecific statistics like min, max, mean and standard deviation are computed as well as domain specific statistics like percentage of cloud, ocean or land pixels, for example.

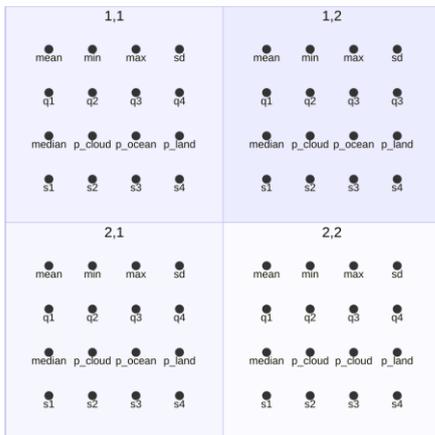

Figure 4: Statistics computed inside grids of HSI

The creation of the index hierarchy at different resolution levels is domain agnostic. But the statistical summary of the underlying data is not, as it can contain domain specific statistics. Based on the computed statistics, the samples in Figure 3 can be selected, e.g., by removing samples with high cloud coverage.

### D. De-biasing of tensors using the Streaming Tensor Sampler

In many domains, some regions of the data corpus are more relevant than others. A practical example is EO imagery. In many cases, imagery from oceans is not as relevant as imagery from land. In addition, depending on the use case, urban or non-urban areas are more relevant. Using HSI, such bias in data can be mitigated. As HSI contain domain specific statistical summaries of underlying discrete and continuous data at different hierarchical resolutions, the consumer can decide how often a tensor of a specific class or category has to be shown (if at all) to the model. Figure 5 displays 500 sampled tensors from urban areas in the United States using the categories of the CropScape dataset [21, 22]. We removed non-urban areas and down-sampled open space and low intensity developed areas (blue and orange) to focus on medium and high intensity developed areas (green and red). As visualized in Figure 6, none of the four sampled categories are in the top 10 most occurring categories of the CropScape dataset. Furthermore, tensors from different data sets can be incorporated by just sampling from different super-tensors, e.g., using different satellite datasets, and via HSI, these datasets can be properly balanced.

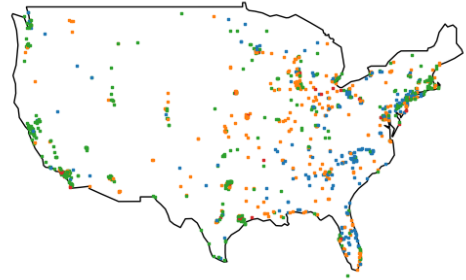

Figure 5: Sampled HSI from urban areas (size not to scale)

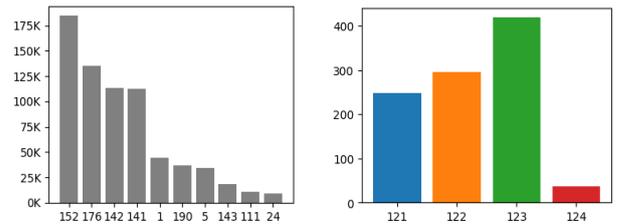

Figure 6: Number of most occurring classes in CropScape (gray) and exemplary number of sampled tensors (colored)

To automate this, we have created the Streaming Tensor Sampler (STS) which consists of a generic PyTorch dataset type [23] and a corresponding dataset factory which takes query parameters as input and returns a parameterized dataset instance. Algorithm 1 below illustrates the pseudo code of the STS. This STS instance can be directly passed to a PyTorch dataloader [23], which takes care of batching, shuffling, parallel processing and data prefetching, among other functions. It is key to notice, that using our stack, no custom line of code needs to be written anymore. The STS directly addresses tensors in the underlying storage system by index. In case of COS, and most other types of file system like storage, this resembles into reading blocks of bytes from a file within a hierarchical file path. Although our stack is also used in production on a key value store, Apache HBase [24] in this case, we won't go into more details here as this backend is neither cloud native [25], nor open-source.

---
**Input:** HSI with labels $L$, column $c$, target ratios $R$, default ratio $r_0$
Initialize pseudo random function $rand$
Initialize list of sampled indices $I_{sampled}$
**For** $l_i$, $r_i$ **in** L, R:
$\quad I_{label}$ = HSI[HSI[c] = $l_i$].index
$\quad I_{sampled}$ = $I_{sampled}$ + rand($I_{label}$, p = $r_i *$ len($I_{label}$))
$I_{label}$ = HSI[HSI[c] not in L].index
$I_{sampled}$ = $I_{sampled}$ + rand($I_{label}$, p = $r_0 *$ len($I_{label}$))
HSI = HSI[$I_{sampled}$]
**Return** HSI

---
Algorithm 1: Pseudo code of the de-biased sampling

## IV. PERFORMANCE AND SCALABILITY

### A. Introduction and setup

As introduced before, we are targeting cloud native deployments with segregated storage. This means, the data transfer bottleneck is always to be searched for in the network connection separating the storage system from the compute nodes. Therefore, we need to test for only two properties: (1) can our system saturate network bandwidth between compute node and storage system, and (2) does our system linearly scale (obviously eventually saturating network switching fabric's capacity)? To test (1) and (2), we have conducted the following experiments:

- Saturation test in HPC data center using an IBM DS8000 [26] GPFS [27] storage system connected with 50 GBit/s
- Saturation test on an AWS hpc7g.16xlarge instance connected to AWS S3 with 25 GBit/s
- Saturation and scaling test on IBM CodeEngine [28], 10 GBit/s per worker to IBM Cloud Object Storage [29]

We are randomly accessing indices on the super-tensor and streaming sub-tensors of 8 MB size back to the client. We increase number of parallel threads until we saturate the network. In case of the scaling experiment, we ran multiple instances of the saturated configuration.

### B. Results

As illustrated in Figure 7, in the HPC data center we could show to saturate the 50 GBit/s link using 10 parallel threads by obtaining a tensor stream rate of 762.5 tensors per second, which corresponds to ~6.1 GB/s. On AWS we saturated the 25 GBit/s link using 128 threads with 387.5 tensors per second, corresponding to ~3.1 GB/s. In Figure 8, we show that significantly more threads are necessary to saturate the storage network connection. Although not yet verified, we assume HTTP imposes some significant overhead and latency in comparison to the protocol used by GPFS. Finally, on CodeEngine (Figure 9) we've saturated the 10 GBit/s link using 64 threads with 137.5 tensors per second, corresponding to ~1.1 GB/s.

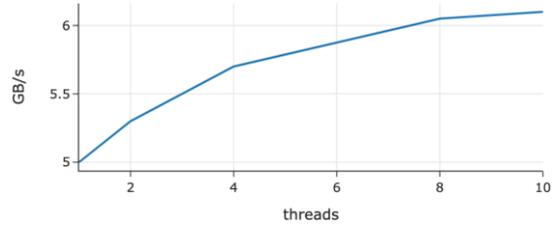

Figure 7: Results from the HPC data center using GPFS on a 50 GBit/s link

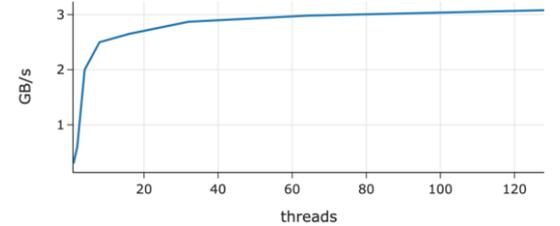

Figure 8: Results on AWS using a hpc7g.16xlarge instance with 25 GB/s link to S3

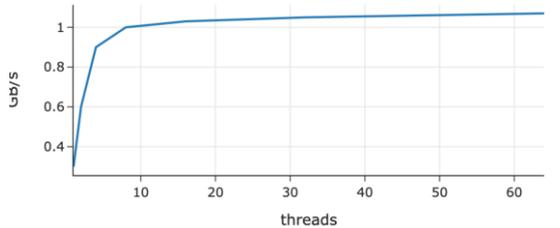

Figure 9: Results on IBM Code Engine using a 12x48 container instance with 10 GB/s link to COS

Finally, we've conducted a scale-out test. Using the saturated instances of CodeEngine, we increased the number of parallel code containers until we saturated. CodeEngine runs on IBM Cloud using bx2-16x64 instances which are virtual machines with 32 GBit/s of network bandwidth, which is shared between network and storage via up to five virtual network interface cards (vNICs). As bx2-16x64 instances have 16 vCPU and 64 GB of memory, we've chosen 12 vCPU/48 GB CodeEngine container size to make sure no two of our containers are co-located on the same virtual machine. We illustrate the results in Figure 10, where we've fixed the number of threads per instance to 64 and started to saturate at roughly 10 GB/s.

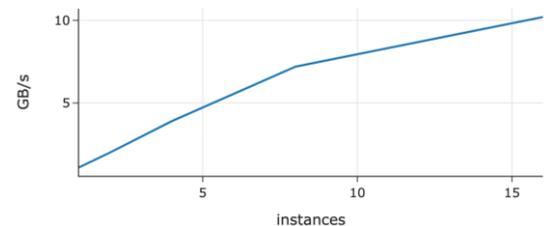

Figure 10: Results on IBM Code Engine using up to 16 12x48 container instances with 10 GB/s link to COS

## V. CONCLUSION

Our TensorBank stack shows significant improvements in ease of use since Analytics Ready Data (ARD) can be directly consumed without additional code to be written. By allowing for efficient filtering and de-biased sampling via HSI, ARD doesn't have to be re-created on a per experiment basis anymore. Thus, TensorBank saves storage costs by avoiding per experiment data transformation and duplication. Also, TensorBank reduces wait time and coordination efforts and waiting time as data preparation is often done by different teams than the data consumers. We've shown that the bare minimum of 1.1 GB/s data ingestion speed to the compute nodes, resulting in triple digit tensor/s rate, is more than enough for our models which usually run minutes to hours per batch on Nvidia A100 GPUs.